\documentclass[1column]{article}
\usepackage[pdftex]{graphicx}
\usepackage[a4paper, total={6in, 8in}]{geometry}
\usepackage[font=footnotesize, margin=1cm]{caption}
\usepackage{float}
\usepackage{subfig}
\usepackage{epstopdf}
\usepackage{color}
\DeclareGraphicsExtensions{.bmp,.png,.pdf}

\usepackage{amssymb}
\usepackage{amsmath}
\usepackage{multirow}
\usepackage[ruled,vlined]{algorithm2e}
\usepackage[]{algorithm2e}

\newcommand{\doublehat}[1]{\Hat{\Hat{#1}}}

\begin{document}
\title{Prediction and Sampling with Local Graph Transforms for Quasi-Lossless Light Field Compression}
\author{Mira~Rizkallah,~Thomas~Maugey,~and~Christine~Guillemot,~\\
        INRIA Rennes Bretagne Atlantique, Rennes, France
\thanks{This work has been supported by the EU H2020 Research and Innovation Programme under grant agreement No 694122 (ERC advanced grant CLIM).}}


\maketitle

\begin{abstract}
Graph-based transforms have been shown to be powerful tools in terms of image energy compaction.  However, when the support increases to best capture signal dependencies, the computation of the basis functions becomes rapidly untractable. 
This problem is in particular compelling for high dimensional imaging data such as light fields.
The use of local transforms with limited supports is a way to cope with this computational difficulty. Unfortunately, the locality of the support may not allow us to fully exploit long term signal dependencies present in both the spatial and angular dimensions in the case of light fields. This paper describes sampling and prediction schemes with local graph-based transforms enabling to efficiently compact the signal energy and exploit dependencies beyond the local graph support. The proposed approach is investigated and is shown to be very efficient in the context of spatio-angular transforms for quasi-lossless compression of light fields. 

\end{abstract}


\section{Introduction}

A graph is a useful tool to describe intrinsic image structures. The graph can then be used as a support for defining and computing de-correlation transforms, which is a critical step in image compression. Fourier-like transforms, called graph Fourier transform (GFT) \cite{shuman2013} and many
variants \cite{shen2010edge,Hu2012,kim2012graph,Zhang2013,hu2015multiresolution,su2017graphfourier} have been shown to be powerful tools for coding piecewise smooth and natural 2D images. 
An interesting review can be found in \cite{cheung}.
However, when the dimension of the signal increases, the dimension of the graph increases and the  complexity inherent to the computation of the GFT basis functions rapidly becomes untractable. This is obviously the case for light field data making a complete graph connecting all light rays unsuitable for this task.  

To cope with this difficulty, we consider instead local transforms with limited supports.
In order to take into account the scene geometry, the support of the graph is defined by super-rays. Super-rays have been first introduced in \cite{Hog2017} as an extension of super-pixels in the 3D domain to group light rays coming from the same 3D object, i.e. to group pixels having similar color values and being close spatially in the 3D space. 
While the locality of the support allows us to reduce the computation complexity of the basis functions, it does not allow us to capture long term spatial dependencies of the signal, unlike efficient predictive schemes used in state of the art codecs (e.g. HEVC). The correlation between different super-rays is not exploited. 

In this paper, we introduce sampling and prediction schemes to exploit correlation beyond the limits of the local graph transform support. 
More precisely,
based on the graph sampling theory, the proposed methods allow taking advantage of the
good energy compaction property of graph transforms on
local supports, i.e. with a limited complexity, while benefiting
from well established but powerful prediction mechanisms
in the pixel domain.
The idea is to first sample the light field data and to encode these references samples with any image coder having powerful Intra prediction mechanisms. The local graph transform is then computed, but only its high frequency coefficients are coded and transmitted. We derive the equations 
allowing us to recover the low frequency coefficients of the local graph transforms from its coded high frequency coefficients and from the encoded reference samples. The encoding of these reference samples is a way to efficiently encode the low frequency coefficients containing most of the light field energy, using Intra prediction mechanisms of state-of-the-art coders. In the experiments, we used HEVC Intra (HM 16.10).

In this general framework, one key question to address is the best choice of the reference samples. The most natural way would be to take all the pixels of a reference view. However, due to matrix conditioning problems, that we will discuss in the paper, the recovered low frequencies are, in that case, very sensitive to high frequencies coefficients quantization. In order to overcome this issue, we sample the graph in each super-ray, across views, and project the samples into one reference image. 
Although this approach gives good performance in terms of energy compaction and quasi-lossless compression of the light field, using a complete graph per super-ray still suffers from complexity limitations and high sensitivity to the quantization noise present in the high frequency coefficients.

To further decrease the basis function computational complexity, we then consider separable local graph transforms applying first a spatial followed by an angular transform. Unlike in the non separable case, the prediction equations do not suffer from numerical instabilities and from the presence of quantization noise in the high frequencies. The reference samples can thus be taken from a light field view.
This second approach keeps the advantages of both the reduced basis function computational complexity due to the limited support and of the structured set of reference samples (one entire view) that can be easily coded with intra-prediction mechanisms. 
It however keeps only in part the advantage of the energy compaction of the graph transform since the recovered frequencies do not necessarily correspond to the low frequencies. This second approach is partly based on the spatio-angular prediction scheme described in \cite{MiraDCC2019} for non separable spatio-angular graph transforms.

The proposed methods can be seen as graph-based prediction schemes deriving low frequency spatio-angular coefficients from one single compressed reference image (e.g. the projected set of reference samples in the non-separable case, or the top-left view in the separable case) and from the high frequency coefficients.
The methods have been assessed in the context of quasi-lossless encoding of light fields. 
Experimental results show that, when coupled with a powerful intra-prediction tool, the graph-based spatio-angular prediction brings a substantial gain in bitrate reaching almost $30 \%$. 

\section{Related work}

In this section, we first review the basics of graph-based transforms and of graph sampling theory. We then give a quick overview of the approaches considered so far for light field compression.

\subsection{Graph transforms}
A graph has been shown to be a useful tool to describe the intrinsic image structure, hence to capture correlation, which is necessary for image compression.
An interesting review  of  graph spectral image processing can be found in \cite{cheung}.

For image compression, 
the signal is defined on an undirected connected graph $\mathcal{G} = \{\mathcal{V},\mathcal{E}\}$ which consists of a finite set $\mathcal{V}$ of vertices corresponding to the pixels. A set $\mathcal{E}$ of edges connect each pixel and its 4-nearest neighbors in the spatial domain. By encoding
pixel similarities into the weights associated to edges, the undirected
graph encodes the image structure.
A Fourier-like transform for graph signals called graph Fourier transform (GFT) \cite{shuman2013} and many
variants \cite{shen2010edge,Hu2012,kim2012graph,Zhang2013,hu2015multiresolution,su2017graphfourier} have been used as adaptive transforms for coding piecewise smooth and natural images. 

A spectrum of graph frequencies can be defined through the eigen-decomposition of the graph Laplacian matrix $\mathbf{L}$ defined as $\mathbf{L = D - A}$, where $\mathbf{D}$ is a diagonal degree matrix whose $i^{th}$ diagonal element $D_{ii}$ is equal to the sum of the weights of all edges incident to node $i$. The matrix $\mathbf{A}$ is the adjacency matrix with 
entries $A_{mn}=1$, if there is an edge $e = (m,n)$ between two vertices $m$ and $n$, and  $A_{mn}=0$ otherwise.

The Laplacian matrix $\mathbf{L}$ is symmetric positive semi-definitive and therefore can be diagonalized as: 
\begin{equation}
\mathbf{L = U^\top \Lambda U }
\end{equation}
where $\mathbf{U}$ is the matrix whose rows are the eigenvectors of the graph Laplacian and $\mathbf{\Lambda}$ is the diagonal matrix whose diagonal elements are the corresponding eigenvalues.
The eigenvectors $\mathbf{U}$ of the Laplacian of the graph are analogous to the Fourier bases in the Euclidean domain and allow representing the signals residing on the graph as a linear combination of eigenfunctions akin to Fourier Analysis \cite{shuman2013}. 
This is known as the Graph Fourier transform. 

\subsection{Graph Sampling}
Let us consider a graph $\mathcal{G} = \{\mathcal{V},\mathcal{E}\}$ made of $N$ vertices associated with a Laplacian $\mathbf{L}$. It has a complete set of eigenvalues $\lambda_{l}, l \in [1,\ N]$ and eigenvectors $\mathbf{u}_{l}, l \in [1,\ N]$. A graph signal is bandlimited and has a bandwidth $\omega = \lambda_{n}$ if it can be expressed as a linear combination of only the first $n$ eigenvectors of $\mathbf{L}$. The space of $\omega$-bandlimited signals is called a Paley-Wiener space and is denoted as $PW(\mathcal{G})_\omega \subset \mathbb{R}$. 

A subset of vertices $S \subset \mathcal{V}$ is a \textit{uniqueness set} \cite{narang2013localized} for signals in $PW(\mathcal{G})_\omega \subset \mathbb{R}$ if $\forall \mathbf{f},\mathbf{g} \in PW_\omega(\mathcal{G}), \mathbf{f}(\mathcal{S}) = \mathbf{g}(\mathcal{S}) \implies  \mathbf{f} = \mathbf{g}$.
It is also shown that $\mathcal{S}$ is a \textit{uniqueness set} for all signals $\mathbf{f} \in PW_\omega(\mathcal{G})$, if and only if  $[\mathbf{u}_1(\mathcal{S}) \mathbf{u}_2(\mathcal{S}), \dots, \mathbf{u}_n(\mathcal{S})]$ are linearly independent where $\lambda_n$ is the $n^{th}$ smallest eigenvalue of $\mathbf{L}$ and $\mathbf{u}_i(S) \in \mathbb{R}^{|S|}$ is a reduced eigenvector. The term reduced implies taking the rows of the eigenvectors corresponding to the indices of the sampling set $\mathcal{S}$ \cite{tzamarias2018novel}. 
It can also be shown that for any minimum uniqueness set $\mathcal{S}$ of size $n$ for signals in $PW_\omega(\mathcal{G})$, there is always at least one node
$\mathcal{s}_i \notin \mathcal{S}$ such that $\mathcal{S} \bigcup \mathcal{s}_i$
is a uniqueness set of size $n + 1$ for
signals in $PW_{\omega+1}(\mathcal{G})$ \cite{tzamarias2018novel}. This property will be useful for iteratively selecting the set of reference samples from the input light field data.

After building a uniqueness set, a simple way to reconstruct the missing samples is to solve a least-squares problem in the spectral domain \cite{narang2013localized}. Observing that the signal $\mathbf{f} \in PW_\omega(\mathcal{G})$ can be written as
\begin{eqnarray}
\mathbf{f}&=&\left[\begin{array}{c}\mathbf{f}(\mathcal{S}) \\ 
\mathbf{f}(\mathcal{S}_c)
\end{array}\right] \\\nonumber
&=& \left[ \begin{array}{cccc}
\mathbf{u}_1(\mathcal{S})& \mathbf{u}_2(\mathcal{S})& \dots& \mathbf{u}_n(\mathcal{S}) \\ \hline
\mathbf{u}_1(\mathcal{S}_c)& \mathbf{u}_2(\mathcal{S}_c)& \dots& \mathbf{u}_n(\mathcal{S}_c) \end{array}\right] \left[\begin{array}{c}
\alpha_1\\
\alpha_2\\
\cdots\\
\alpha_n
\end{array}\right],
\end{eqnarray}

the vector $\left[\alpha_1, \alpha_2, \dots,\alpha_n\right]^\top$ can be retrieved by searching for the least square solution of the upper part of the above system as:
\begin{equation}
\left[\alpha_1, \alpha_2, \dots,\alpha_n\right]^\top =  \left(\tilde{\mathbf{U}}^\top(\mathcal{S}) \tilde{\mathbf{U}}(\mathcal{S})\right)^{-1}\tilde{\mathbf{U}}(\mathcal{S})\mathbf{f}(\mathcal{S}), 
\label{eq:LSreconstruction}
\end{equation}
then the missing samples are reconstructed as follows: 
\begin{equation}
\mathbf{f}(\mathcal{S}_c) = \tilde{\mathbf{U}}(\mathcal{S}_c) \left[\begin{array}{c}
\alpha_1\\
\alpha_2\\
\cdots\\
\alpha_n
\end{array}\right]
\label{eq:LSreconstruction_2}
\end{equation}
where columns of $\tilde{\mathbf{U}}$ are the $n$ first eigenvectors of the $\mathbf{L}$. 

In the special case where $\mathcal{S}$ is of size $n$ ($\mathcal{S}$ is therefore a \textit{minimum uniqueness set} \cite{tzamarias2018novel} for signals $\mathbf{f} \in PW_\omega(\mathcal{G})$), $\tilde{\mathbf{U}}(\mathcal{S})$ is a square invertible matrix. 
Equipped with the aforesaid arguments, the formulation in Equation \ref{eq:LSreconstruction} can be further simplified to:
\begin{equation}
\left[\alpha_1, \alpha_2, \dots,\alpha_n\right]^\top =  \left( \tilde{\mathbf{U}}(\mathcal{S})\right)^{-1}\mathbf{f}(\mathcal{S}), 
\label{eq:LSreconstruction_simplified}
\end{equation}

While the aforementioned sampling theorem \cite{narang2013localized} has been proposed for band-limited signals, we extend those equations to our problem in the following section. More precisely, we deal with signals (i.e. Color Signals) that might not be necessarily band-limited on the underlying graph supports (i.e. Super-Rays).

\subsection{Light field compression}
The availability of commercial light field cameras has given momentum to the development of light field compression algorithms. 
Many solutions proposed so far adapt standardized image and video compression solutions (in particular HEVC) to light field data. This is the case \textit{e.g.} in \cite{conti2012new, conti2016hevc,conti2016hevcvideo,li2016compression}, 
where the authors extend HEVC intra coding modes by adding new prediction modes to exploit similarity between lenslet images. This is also the case in \cite{liu2016pseudo,rizkallah2016impact,Jia2017}, 
where the views are encoded as pseudo video sequences using HEVC or the latest JEM software, or in \cite{Ahmad2017} where HEVC is extended for coding an array of views.


Low rank models as well as local Gaussian mixture models in the 4D rays space are proposed in \cite{jiang2017light}, \cite{ElianDCC2019} and \cite{verhack2017steered} respectively. View synthesis based predictive coding has also been investigated in \cite{zhao2017light} where the authors use a linear approximation computed with Matching Pursuit
View synthesis based predictive coding is another research direction followed in \cite{zhao2017light} where the authors use a linear approximation computed with Matching Pursuit
for disparity based view prediction. The authors in \cite{jiang2017hot3D} and \cite{su2017graph} use instead a the convolutional neural network (CNN) architecture proposed in \cite{kalantari2016learning} for view synthesis and prediction. The prediction residue is then coded using HEVC \cite{jiang2017hot3D}, or using local residue transforms (SA-DCT) and coding \cite{su2017graph}. 
The authors in \cite{tabuslossy}, use a depth based segmentation of the light field into 4D spatio‐angular blocks with prediction followed by JPEG‐2000.
View synthesis followed by predictive coding is the approach followed in JPEG-Pleno \cite{plenovm11}. While all prior work mentioned above has been dedicated to lossy compression, much less effort has been dedicated to lossless coding of light fields. One can however mention the approach proposed in \cite{perra2015lossless} using differential prediction.

\section{Super-ray based graph transforms}

Let us consider the 4D representation of light fields proposed in \cite{LEV96} and \cite{GOR96} describing the radiance along rays by a function $L(s,t,m,n)$. Based on this representation, the light field of dimensions $(S,T,M,N)$ can be regarded as an array of views at angular positions $(m,n)$, each view being composed of pixels with spatial coordinates $(s,t)$. In the sequel, to denote one view, we will use the pair of indices $(m,n)$ or an index $v=(m,n)$ to simplify the notations. 

Such light fields represent very large volumes of high dimensional data. Graphs connecting all light rays spatially and across views can rapidly become untractable, and in particular the computation of the basis functions, if we consider one unique graph for the entire light field. To overcome this difficulty, here we consider graphs with limited supports defined by super-rays in order to follow the scene geometry. 

\subsection{Super-ray construction}
\label{sec:GraphConstruction}
The concept of super-ray has been initially introduced in \cite{Hog2017} as an extension of super-pixels
to address the computational complexity issue in light field image processing tasks. 
The term {\em super-pixel}, first coined in \cite{ren2003learning} can be seen as the clustering of image pixels into a set of perceptually uniform regions. Similarly, super-rays can be seen as the clustering of rays of the light field within and across views, hence corresponding to the same set of 3D points of the imaged scene. 

\begin{figure}[ht!]
	\centering
	\includegraphics[width=0.48\textwidth]{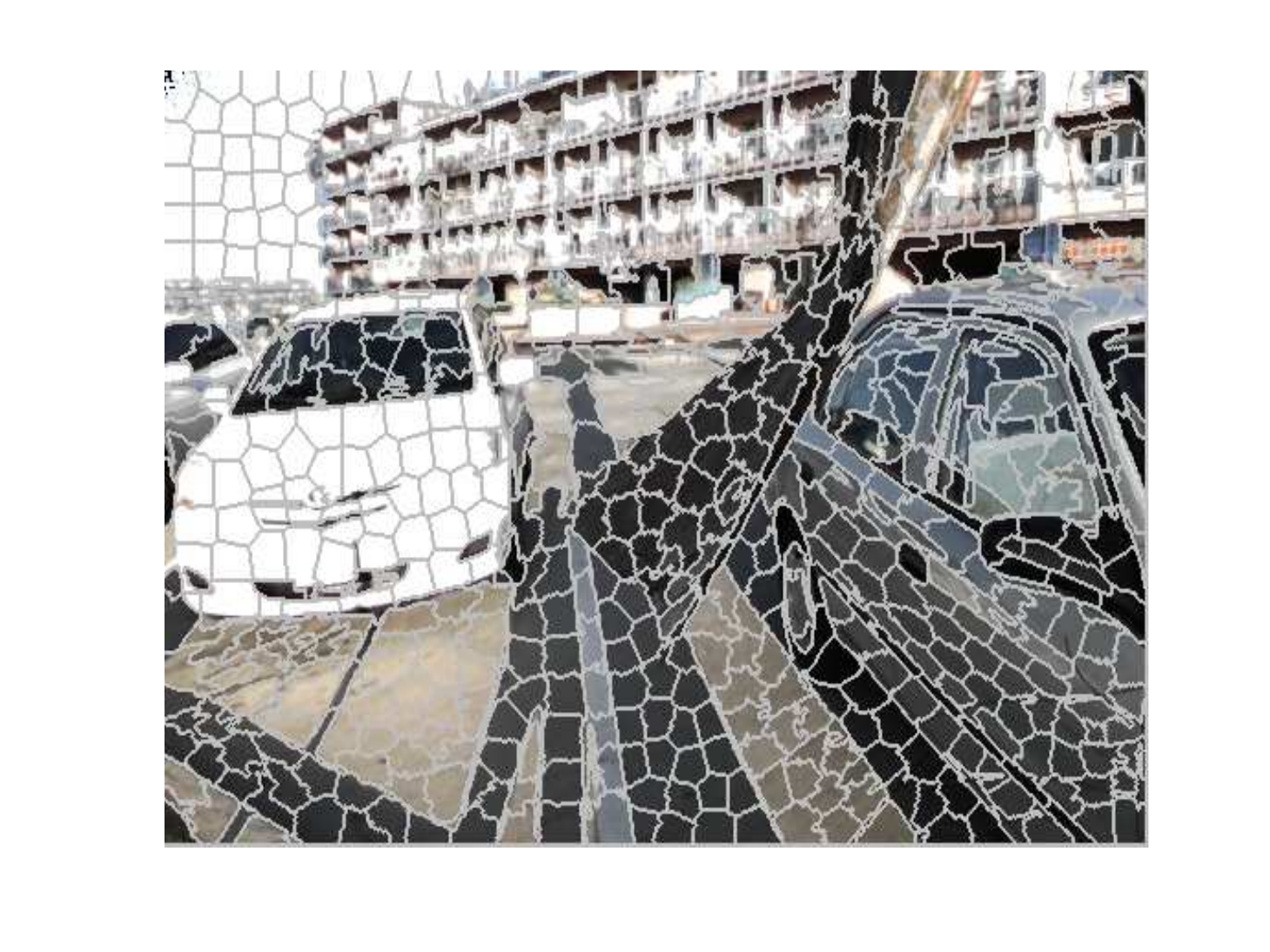}
	\caption{An example of super-pixel segmentation for the \textit{Cars} dataset, obtained by fixing the number of superpixels to $800$.}
	\label{fig:original_seg}
\end{figure}

To construct super-rays, we proceed as follows. 
We first compute super-pixels in the top-left view using the SLIC algorithm \cite{achanta2012slic} as well as its disparity map using the method in \cite{jiang2018depth}. An example of an original image with its segmentation is shown in Fig \ref{fig:original_seg}. 

Then, using the disparity map, we compute the median disparity per super-pixel and use this median disparity to project the segmentation labels to all the other views. 
The algorithm proceeds row by row. In the first row of views, we perform horizontal projections from the top-left $I_{1,1}$ to the $N-1$ views next to it. For each other row of views, a vertical projection is first carried out from the top view $I_{1,1}$ to recover the segmentation on view $I_{m,1}$, then $N-1$ horizontal projections from $I_{m,1}$ to the $N-1$ other views are performed. At the end of each projection, some labels are projected in all the views without interfering with others. Those typically represent flat regions inside objects. Others mainly consisting of occluded and occluding segments end up superposed in some views. In this case, the occluded pixels are assigned the label of the neighboring super-ray corresponding to the foreground objects (\textit{i.e.} having the higher disparity). As for appearing pixels, they are clustered with the background super-rays (\textit{i.e.} having the lower disparity.

\begin{figure*}[ht!]
	\centering
	\includegraphics[width=0.95\linewidth]{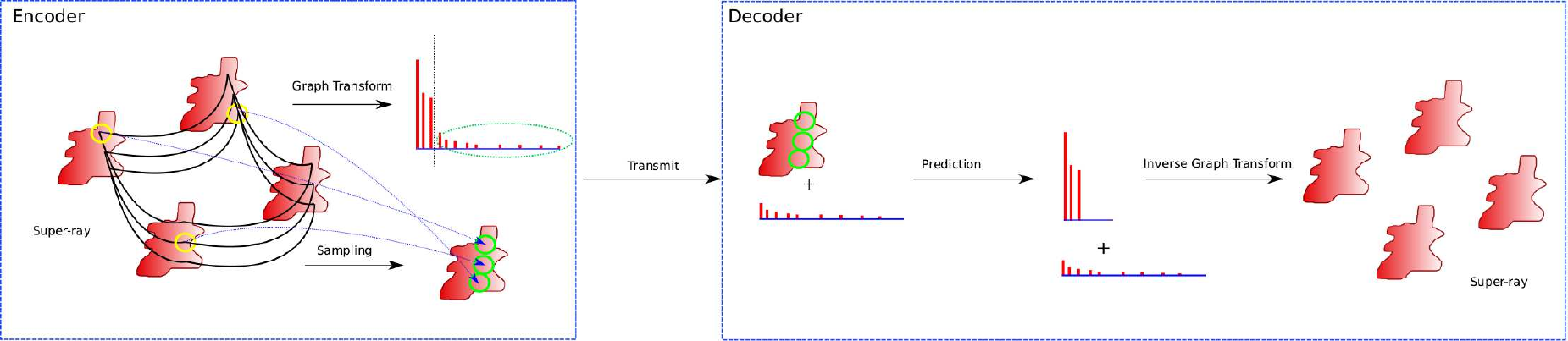}\\
	\caption{Overview of the prediction and sampling scheme with a local graph transform applied on a super-ray. On the encoder side, a sampling is performed to find the best set of samples and a graph transform is applied. The high frequencies along with the reference samples (surrounded in green) are sent to the decoder which predicts the low frequency transform coefficients to then recover the original super-ray by inverse graph transform.}
	\label{fig:methodology_example}
\end{figure*}

\subsection{Local graph transforms}

We will denote the luminance values of all the light rays (i.e. pixels across all the views) in the $k^{th}$ super-ray, by the vector $\mathbf{x}_k \in \mathcal{R}^{N_k}$, where $N_k$ is the number of rays in the $k^{th}$ super-ray. The $k^{th}$ super-ray $\mathbf{x}_k$ is formed by a set of super-pixels (corresponding super-pixels across the different views). Each super-pixel forming the $k^{th}$ super-ray $\mathbf{x}_k$ will be denoted; in a vectorized form,
$\mathbf{x}_{k,v} \in \mathcal{R}^{N_{k,v}}$. 

We build a \textit{4-connected graph} inside each super-ray i.e. each pixel is connected to its 4 nearest neighbors (horizontal and vertical neighbors). The graph transform of the $k^{th}$ super-ray $\mathbf{x}_k$ is defined as
\begin{equation}
\hat{\mathbf{x}}_k = \mathbf{U}_k^\top \mathbf{x}_k
\end{equation}
Where the columns of $\mathbf{U}_k$ are the eigenvectors of the local graph laplacian inside the $k^{th}$ super-ray.
The inverse graph Fourier transform is then given by
\begin{equation}
\mathbf{x}_k = \mathbf{U}_k \hat{\mathbf{x}}_k
\end{equation}

However, computing the transform on a local support does not allow us to exploit spatial signal dependencies outside the support, resulting in some loss in compression efficiency. To exploit these dependencies, some form of prediction across super-rays would be needed. Nevertheless, the super-rays being of arbitrary shapes, developing inter super-ray prediction mechanisms is not an easy task. The idea we develop here consists instead in encoding a selected set of samples, using powerful prediction mechanisms available in state-of-the-art coders (e.g. HEVC), and then to  recover the low frequency coefficients of
the local graph transforms from its coded high frequency coefficients and the encoded reference samples as seen in Fig \ref{fig:methodology_example}. 

\section{Super-ray based graph prediction and sampling}

\subsection{Graph-based prediction}


%


Let us denote $\mathcal{S}_k$ the set of pixel indices in a super-ray $k$ and 
$\mathcal{S}$ those belonging to the sampling set. Let $\mathcal{S}_C = \mathcal{S}_k \setminus \mathcal{S} $ be 
the set of all other pixels indices of the super-ray. Let $N_\mathcal{S}$ be the cardinal of $\mathcal{S}$.
We denote $\mathcal{T}$  the set of $N_\mathcal{S}$ lowest frequency coefficients and $\mathcal{T}_C$ the rest of the frequency coefficients.

Due to the high level of correlation between the different pixels forming a super-ray $k$, the energy of the transformed coefficients $\hat{\mathbf{x}}_k$ is highly compacted in the low frequencies $\hat{\mathbf{x}}_k(\mathcal{T})$. However, we might still end up with some non-zero high frequencies $\hat{\mathbf{x}}_k(\mathcal{T}_c)$. 
If we choose an appropriate uniqueness sampling set $\mathcal{S}$ in the $k^{th}$ super-ray, then 
the inverse graph transform is defined under appropriate permutation as
\begin{equation}
\mathbf{x}_k = \mathbf{U}_k \hat{\mathbf{x}}_k 
\end{equation}
i.e., as
\begin{equation}
\left[ \begin{array}{c} \mathbf{x}_k(\mathcal{S}) \\ \hline 
\mathbf{x}_k(\mathcal{S}_c)
\end{array} \right] = \left[ \begin{array}{c|c}
\mathbf{U}_k(\mathcal{S},\mathcal{T})  & \mathbf{U}_k(\mathcal{S},\mathcal{T}_c)  \\  \hline \mathbf{U}_k(\mathcal{S}_c,\mathcal{T}) &   \mathbf{U}_k(\mathcal{S}_c,\mathcal{T}_c) \end{array}  \right] \left[\begin{array}{c}
\hat{\mathbf{x}}_k(\mathcal{T})\\ 
\hline \hat{\mathbf{x}}_k(\mathcal{T}_c)
\end{array} \right].
\end{equation}
If the signal samples are transmitted separately, $\mathbf{x}_k(\mathcal{S})$ is available at the decoder. If we impose $|\mathcal{S}| = |\mathcal{T}|$, then  $\mathbf{U}_k(\mathcal{S},\mathcal{T})$ is a square invertible matrix. Furthermore, if we only transmit $\hat{\mathbf{x}}_k(\mathcal{T}_c)$, then we are able to recover $\hat{\mathbf{x}}_k(\mathcal{T})$ from the following equation:
\begin{equation}
\hat{\mathbf{x}}_k(\mathcal{T}) = \Big(\mathbf{U}_k(\mathcal{S},\mathcal{T})\Big)^{-1} \Big(\mathbf{x}_k(\mathcal{S}) -  \mathbf{U}_k(\mathcal{S},\mathcal{T}_c) \hat{\mathbf{x}}_k(\mathcal{T}_c) \Big). \label{eq:predictionNS}
\end{equation}

Equation (\ref{eq:predictionNS}) is our so-called graph-based spatio-angular prediction. First, $\mathbf{x}_k(\mathcal{S})$ can be seen as a signal composed of a $\lambda_{|\mathcal{S}|}$-band-limited part plus some high frequencies. In this equation, we are actually removing the high frequencies to retrieve the band-limited signal (i.e. $\mathbf{x}_k(\mathcal{S}) -  \mathbf{U}_k(\mathcal{S},\mathcal{T}_c) \hat{\mathbf{x}}_k(\mathcal{T}_c)$). Using the least squares reconstruction method in (\ref{eq:LSreconstruction}), we find the low frequency transformed coefficients $\hat{\mathbf{x}}_k(\mathcal{T})$.

Moreover, the high-frequency coefficients $\hat{\mathbf{x}}_k(\mathcal{T}_c)$ can be also seen as prediction coefficients, transmitted to recover the exact light field at the decoder. The basis of the linear prediction is the graph-transform basis, which makes these coefficients low-energetical and thus easy to transmit.

The signal values at $\mathcal{S}_c$ are then retrieved as
\begin{equation*}
\hat{\mathbf{x}}_k(\mathcal{S}_c) =
\mathbf{U}_k(\mathcal{S}_c,\mathcal{T})  \hat{\mathbf{x}}_k(\mathcal{T}) +  \mathbf{U}_k(\mathcal{S}_c,\mathcal{T}_c) \hat{\mathbf{x}}_k(\mathcal{T}_c),
\end{equation*}
where the first term is equivalent to the $\lambda_{|\mathcal{S}|}$-band-limited signal recovered on $\mathcal{S}_c$ and the second term is added in order to take into account the high frequency components.

To be able to carry out our graph-based spatio-angular prediction, we should at first determine the appropriate sampling set. More precisely, we want to find $\mathcal{S}$ that results in the best conditioning of the sub-matrix $\mathbf{U}_k(\mathcal{S},\mathcal{T})$ that guarantees a small reconstruction error. Simultaneously, we seek a sampling set $\mathcal{S}$ that can be wrapped onto one single view to be coded with efficient prediction mechanisms. 

\subsection{Graph sampling for light fields} 
\label{sec:SamplingSet}
A first intuitive way to define the sampling set per super-ray would be to choose the set of $N_{k,v_0}$ pixels that reside in the reference view $v_0$ that can be subsequently coded with intra HEVC. In our experiments however, we have found that the resulting sub-matrix $\mathbf{U}_k(\mathcal{S},\mathcal{T})$ is ill-conditioned for non-consistent super-rays.
We choose instead to use an adapted version of the algorithm described in \cite{tzamarias2018novel}.

More precisely, for each super-ray $k$, we specify the band-limit frequency as $\lambda^k_n$ with $n_k = N_{k,v_0}$.
We seek to find the optimal sampling set $\mathcal{S}$ that guarantees the exact reconstruction of any signal in $PW(\mathcal{G}_k)_{\lambda^k_n}$. We know that we have a correspondence between the size of the minimum uniqueness set and the signal bandwidth. We therefore want to find a set of $N_{k,v_0}$ samples. In order to find the vertices that belong to this set, we have to find $N_{k,v_0}$ linearly independent rows from the matrix $\tilde{\mathbf{U}}$. We follow the same reasoning as in \cite{tzamarias2018novel} but with slightly different constraints to adapt it to our coding problem. In summary, the algorithm takes as input the graphs of all super-rays and the number of samples per super-ray. At the output of this stage, we want to wrap all the samples in a reference view to be efficiently coded with HEVC. 

While this method allows an optimal sampling per super-ray, yet, it does not guarantee that the output vector is well structured. It is impossible to say that the samples of neighboring super-rays will be efficiently de-correlated using intra-prediction mechanisms of any efficient coder. In extreme cases, we might end up with noisy samples that are very difficult to code. We thus propose to wrap our samples into one reference view taking into account the geometrical information given by our local graph. 

We first observe that our graph laplacian is a sum of two laplacians: The first one includes the connections $\mathbf{L}^s_{k}$ (s for spatial) inside views, and the other $\mathbf{L}^a_{k}$ (a for angular) made of edges between pixels inside different views. $\mathbf{L}^a_{k}$ is actually composed of various connected components, each one corresponding to a 3D point in the scene.
\begin{algorithm}
	\SetAlgoLined
	\KwData{The set of graphs for all super-rays, Segmentation map of a reference view, the sampling set size per super-ray: $\{\mathcal{G}_k = \{\mathcal{V}^k,\mathbf{L}_k\}\}$, $\mathbf{SM}_{ref}$,$\{n_k\}$}
	\KwResult{A reference image made of samples drawn in all super-rays: $\mathbf{I}_{ref}$}
	\ForEach{Super-ray $k$}{
		\textbf{Initialize}: $\mathcal{S} \leftarrow \mathcal{V}^k_i$ where $\mathcal{V}^k_i$ is the vertice corresponding to the centroid of the super-pixel residing in the reference view \;
		Compute $\tilde{\mathbf{U}}$\;
		\For{$m = 2 \to n_k$ }{
			Define $\mathcal{T} = [1, m]$\;
			Compute $z = null(\tilde{\mathbf{U}}(\mathcal{S},\mathcal{T}))$\;
			Normalize rows of $\tilde{\mathbf{U}}(\mathcal{S}_c,\mathcal{T})$\;
			Compute $b = \tilde{\mathbf{U}}(\mathcal{S}_c,\mathcal{T})z$\;
			$i \leftarrow argmax_i(|b(i)|)$\;
			$\mathcal{S} \leftarrow \mathcal{S} \cup \mathcal{S}_c(i)$
		}
		Fill $\mathbf{I}_{ref}$ at the right positions : $\mathbf{I}_{ref}(\mathbf{SM}_{ref} = k) = \mathbf{x}_k(\mathcal{S})$\;
		
	}
	\caption{Light Field Super-ray Graph based Sampling Algorithm}
	\label{alg:SRsampling}
\end{algorithm}

\begin{figure*}[ht!]
	\centering
	\includegraphics[width=0.8\linewidth]{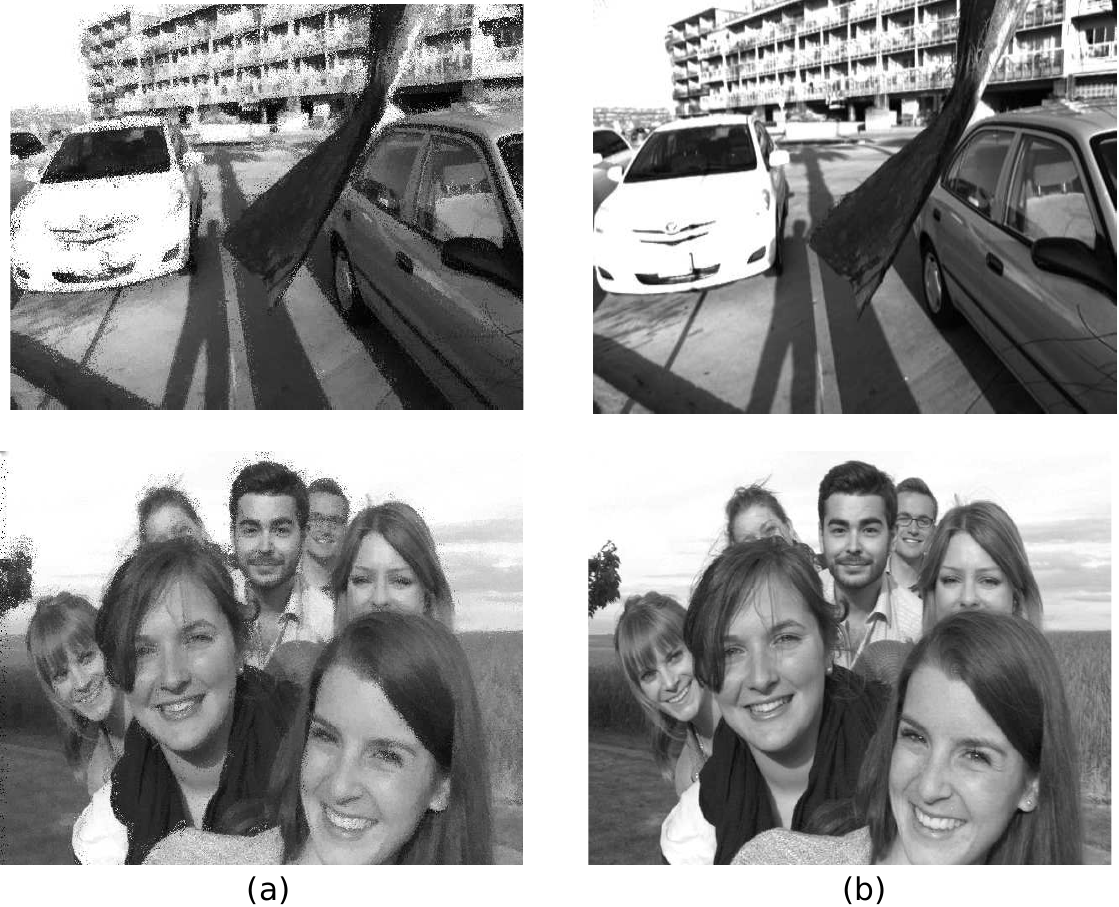}\\
	\caption{Reference images for \textit{Cars} and \textit{Friends}: (a) obtained after projection of the sampling sets in all super-rays in the case of non separable graph transforms; (b) original top-left views used as the reference sampling set in the case of separable graph transforms.}
	\label{fig:samples_example}
\end{figure*}

Using the angular information provided by $\mathbf{L}^a_{k}$, we define the matrix $\mathbf{E}$ of size $(N_{k,v_0} \times N_{k})$ where each element gives the correspondence between a pixel in a super-ray $k$ in $v_0$, and any other pixel in the super-ray. Consider a pixel $p_1$ in the view $v_0$. If we can access a pixel $p_2$ from $p_1$ following the graph connections in $\mathbf{L}^a_{k}$ then the entry $\mathbf{E}(p_1,p_2) = 1$, otherwise $\mathbf{E}(p_1,p_2) = 0$.

For each sample $\mathcal{S}(i)$ corresponding to a point $p$, we find the corresponding point $p_0$ in the set of pixels $\mathcal{S}_0$ belonging to the super-ray in the first view i.e. $p_0$ such as $\mathbf{E}(p,p_0) = 1$. The best case scenario is when each sample has a correspondence to a different pixel in the first view. In this case, the projection is easy following the graph links. In the worst case, more than one sample might have a correspondence with the same point in the first view. In this case, first found, first served. The others are considered as disocclusions, and the pixels having no correspondence in the first view, will be projected into remaining available positions. 
The complete algorithm is summarized in Algorithm \ref{alg:SRsampling}.

Examples of images obtained after the sampling and projection of the selected sample set on a reference view (considering here the Luminance pixel values) are shown in Figure \ref{fig:samples_example}. Despite the non-optimality of this method, we have ascertained that the selected set of reference samples can be efficiently compressed with HEVC using lossless compression settings. Once we have the samples in hand, they can be sent as prediction information to the decoder side, instead of sending the low frequency coefficients. This strategy is a way to efficiently compress the low frequency graph transform coefficients containing about 99 $\%$  of the light field energy, as shown in Table \ref{tab:percentage_energy_predicted} (left column). 

\section{Prediction based on spatio-angular separable graph transforms}
\label{sec:SGT}

To further decrease the basis function computational complexity,
we now consider the case of a separable spatio-angular transform, i.e., applying first a spatial followed by an angular transform, where the set of reference samples is one reference view. We will see that, in addition to a reduced complexity, unlike in the non separable case, the prediction equations do
not suffer from numerical instabilities and from the presence
of quantization noise in the high frequencies.

\subsection{Separable spatio-angular graph transform}
The spatial graph transform coefficients $\hat{\mathbf{x}}_{k,v}$ for each spatial graph $\mathcal{G}_{k,v}$ are obtained by calculating:
\begin{equation}
\hat{\mathbf{x}}_{k,v} = \mathbf{U}_{k,v}^\top \mathbf{x}_{k,v}.
\end{equation}
where $\mathbf{U}_{k,v}$ are the eigenvectors of the spatial laplacian and $\mathbf{x}_{k,v}$ are the luminance values of the super-ray $k$ in view $v$.
Inversely, the luminance values of the pixels belonging to the graph are retrieved from
\begin{equation}
\mathbf{x}_{k,v}= \mathbf{U}_{k,v} \hat{\mathbf{x}}_{k,v} .
\end{equation}

For each super-ray $k$, an angular transform is then used to tract  similarities between the transformed coefficients $\hat{\mathbf{x}}_{k,v}(b)$ of each band $b$ of the spatial transform coefficients $\hat{\mathbf{x}}_{k,v}$, across the views $v$. 
For that, the spatial-band vector is denoted $\hat{\mathbf{x}}^b_k = [\hat{\mathbf{x}}_{k,v}(b)]_{v \in \{1,2,\dots, M \times N\}, \ \mbox{s.t.} \ b < |\mathbf{x}_{k,v}|}$, where $M \times N$ is the total number of views. The angular transform coefficients are then obtained by calculating
\begin{equation}
\doublehat{\mathbf{x}}^b_k = {\mathbf{V}_k^b}^\top \hat{\mathbf{x}}^b_k.
\end{equation}
where $\mathbf{V}_k^b$ is a matrix whose columns are the eigenvectors of the laplacian $\mathbf{L}^b_k$ of the angular graph for the band $b$.

\subsection{Separable graph-based spatio-angular prediction} 
\label{sec:GBSAP}

\begin{figure*}[htp!]
\centering
\includegraphics[width=0.9\textwidth]{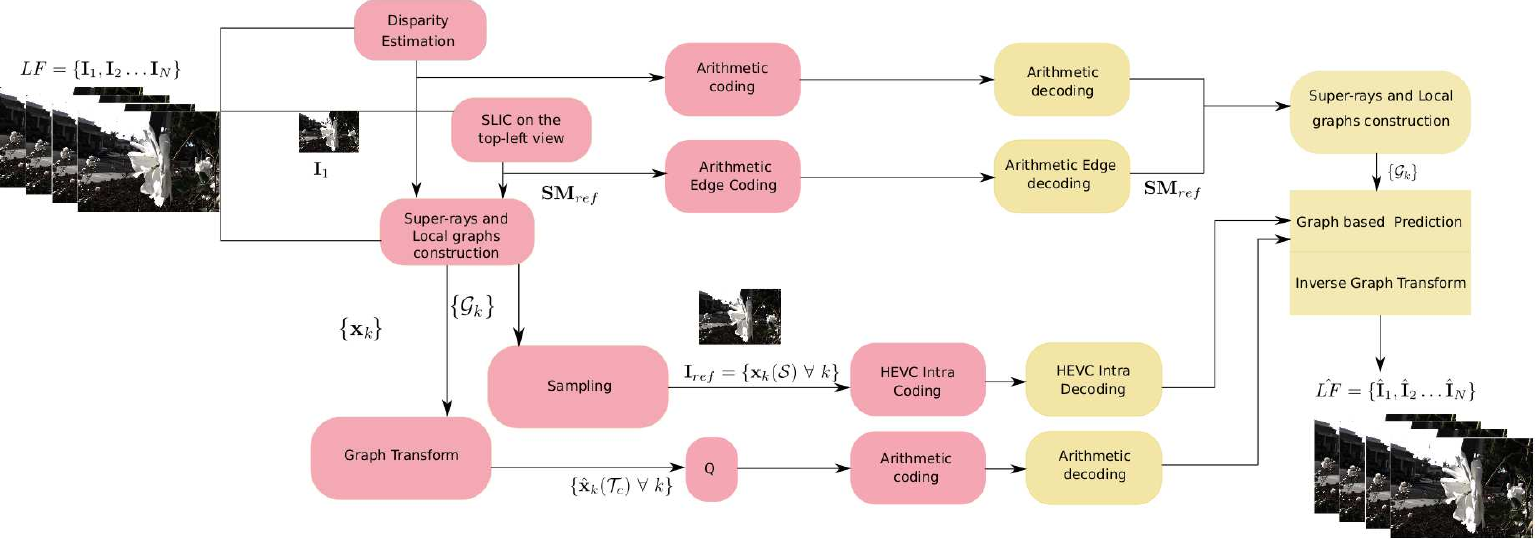}\\
\caption{Overview of proposed coding scheme with the non separable graph transform.}
\label{fig:CodingSchemeNonSeparable}
\end{figure*}
Let us assume that view $1$ is coded as a reference. In order to perform the prediction, we follow the same reasoning as in the previous (non separable graph) case but we apply it to each band that exists in view $1$. 
For a given super-ray $k$, the spatial transform in view $1$ is $\hat{\mathbf{x}}_{k,1} = \mathbf{U}_{k,1}^\top \mathbf{x}_{k,1}$ according to notations introduced before. 

We choose one sample for each band. It corresponds to the vertex $\mathcal{V}_i$ that is in the reference view (labeled by $1$ in our case). For a given band $b$, the inverse angular transform is defined as 
\begin{equation}
\hat{\mathbf{x}}^b_k = \mathbf{V}_k^b \doublehat{\mathbf{x}}^b_k
\end{equation}
i.e., as
\footnotesize
\begin{eqnarray*}
\left[\begin{array}{c} \hat{\mathbf{x}}^b_k(1)\\\hline 
\hat{\mathbf{x}}^b_k(2)\\\vdots\\ \hat{\mathbf{x}}^b_k(N_b)
\end{array}\right]
&=& 
\left[\begin{array}{c|ccc}
\mathbf{V}_k^b(1,1) & \mathbf{V}_k^b(1,2) & \cdots & \mathbf{V}_k^b(1,N_b)\\\hline \mathbf{V}_k^b(2,1) & \mathbf{V}_k^b(2,2) & \cdots & \mathbf{V}_k^b(2,N_b) \\ \vdots & \vdots & \ddots & \vdots\\ \mathbf{V}_k^b(N_b,1)&\mathbf{V}_k^b(N_b,2)&\cdots &\mathbf{V}_k^b(N_b,N_b)\end{array}  \right] 
\\
& \times & \left[\begin{array}{c}
\doublehat{\mathbf{x}}^b_k(1)\\ 
\hline \doublehat{\mathbf{x}}^b_k(2) \\ 
\vdots \\ 
\doublehat{\mathbf{x}}^b_k(N_b)
\end{array} \right]
\label{eq:Separable_prediction_1_big}
\end{eqnarray*}
\normalsize
where $N_b$ denotes the number of views where the $b^{th}$ band of the $k^{th}$ super-ray is defined. 
Since the view $1$ is transmitted separately,  $\hat{\mathbf{x}}^b_k(1)$ is available at the decoder. If we only transmit $\doublehat{\mathbf{x}}^b_k(2), \ldots, \doublehat{\mathbf{x}}^b_k(N_b)$, the we are able to retrieve $\doublehat{\mathbf{x}}^b_k(1)$ from the following equation:


\begin{align}
\begin{split}
&\doublehat{\mathbf{x}}^b_k(1) =
\frac{1}{\mathbf{V}_k^b(1,1)} \times
\\
&\left(\hat{\mathbf{x}}^b_k(1) 
- \left[\begin{array}{ccc} \mathbf{V}_k^b(1,2)&\cdots& \mathbf{V}_k^b(1,N_b) \end{array}\right]
\left[\begin{array}{c} \doublehat{\mathbf{x}}^b_k(2) \\ \vdots \\ \doublehat{\mathbf{x}}^b_k(N_b) \end{array}\right]\right) 
\end{split}
\label{eq:prediction}
\end{align}

Equation (\ref{eq:prediction}) is our graph-based spatio-angular prediction for the separable case. 
The spatial coefficients of all the views are then retrieved from the following equation

\begin{align}
\begin{split}
&\left[\begin{array}{c}
\hat{\mathbf{x}}^b_k(2) \\ \vdots \\ \hat{\mathbf{x}}^b_k(N_b) \end{array} \right] = \left[\begin{array}{c}
\mathbf{V}_k^b(2,1) \\ \vdots \\ \mathbf{V}_k^b(N_b,1) \end{array} \right] \doublehat{\mathbf{x}}^b_k(1)
\\
& \ \ \ + \left[\begin{array}{ccc}
\mathbf{V}_k^b(2,2)&  \cdots & \mathbf{V}_k^b(2,N_b) \\
\vdots & \ddots & \vdots\\
\mathbf{V}_k^b(N_b,2)&\cdots &\mathbf{V}_k^b(N_b,N_b) \end{array} \right] \left[\begin{array}{c}
\doublehat{\mathbf{x}}^b_k(2) \\ \vdots \\\doublehat{\mathbf{x}}^b_k(N_b) 
\end{array}\right] 
\end{split}
\end{align}

Once the decoder has recovered the first spatial graph transform coefficients in all the views, it can reconstruct the whole light field by applying a simple spatial inverse GFT since it has access to the graph supports and coefficients.

\section{Proposed coding schemes}
\label{Codingschemes}
\subsection{Coding scheme with the non separable graph transform} 
Fig.\ref{fig:CodingSchemeNonSeparable} gives an overview of the coding scheme for the non separable case. 
The top left view $\mathbf{I}_1$ is separated into uniform regions using the SLIC algorithm (\cite{achanta2012slic}) to segment the image into super-pixels, and its disparity map is estimated with \cite{jiang2018depth}. The disparity values are encoded using simple arithmetic coder. The segmentation is coded with edge arithmetic coder(AEC) \cite{daribo2012arithmetic}. Using both the segmentation map and the geometrical (disparity) information, we can build consistent super-rays and graphs in and across all views, as explained above, at both the encoder and decoder sides.
Once the local graphs are computed, we can find the optimal sampling sets (their actual positions in the light field and the corresponding luminance values) as explained in \ref{sec:SamplingSet}. Those samples are reorganized in a reference image coded with HEVC intra and sent as prediction information to the decoder.

We apply the non separable graph transform on the coded version of the reference image (quasi-lossless coding) and the original values of all other samples to compact their energy in fewer coefficients. Since the reference image is coded with very small QP, we are almost sure that we are not adding angular incoherence between the different views. Once we have the graph transform coefficients, instead of sending the whole spectrum with simple arithmetic coding, we propose to use the proposed graph-based prediction to derive the low frequency spatio-angular coefficients from the coded reference set of samples and high angular frequency coefficients, at the decoder side. 

We thus send, for all super-rays, the AC coefficients, i.e., the ($N_{k} - N_{k,1}$) last bands obtained with the non separable graph transform. ($N_{k}$ and $N_{k,1}$ are the number of pixels belonging to the super-ray $k$ and those only residing in view $1$ respectively). 
Specifically, after applying the spatio-angular graph transforms on all super-rays, all frequency coefficients are grouped into a two-dimensional array $\mathbf{y}$ where $\mathbf{y}(k,v)$ is the $v^{th}$ transformed coefficient for the super-ray $k$. Using the natural scanning order (increasing order of eigenvalues), we assign a class number to each super-ray. For a class $i$, the high frequencies are defined as the last $round(N_k \times (4-i)/4)$ coefficients where $N_k$ is the total number of coefficients. Each super-ray belongs to class $i$ if it does not belong to class $i-1$ and the mean energy per high frequency is less than $1$. 
More precisely, we start by finding the super-rays in the first class then remove them from the search space before finding the other classes, and similarly for the following steps. We code a flag with an arithmetic coder to give the information of the class of super-rays to the decoder side. In class $i$, the last  $round(N_k \times (4-i)/4)$ coefficients of each super-ray are discarded. The remaining  high frequency spatio-angular coefficients are quantized uniformly with a small step size $Q=0.5$. They are grouped into $32$ uniform groups to be arithmetically encoded. 

\subsection{Coding scheme with the separable graph Transform}

\begin{figure*}[ht!]
\centering
\includegraphics[width=0.9\textwidth]{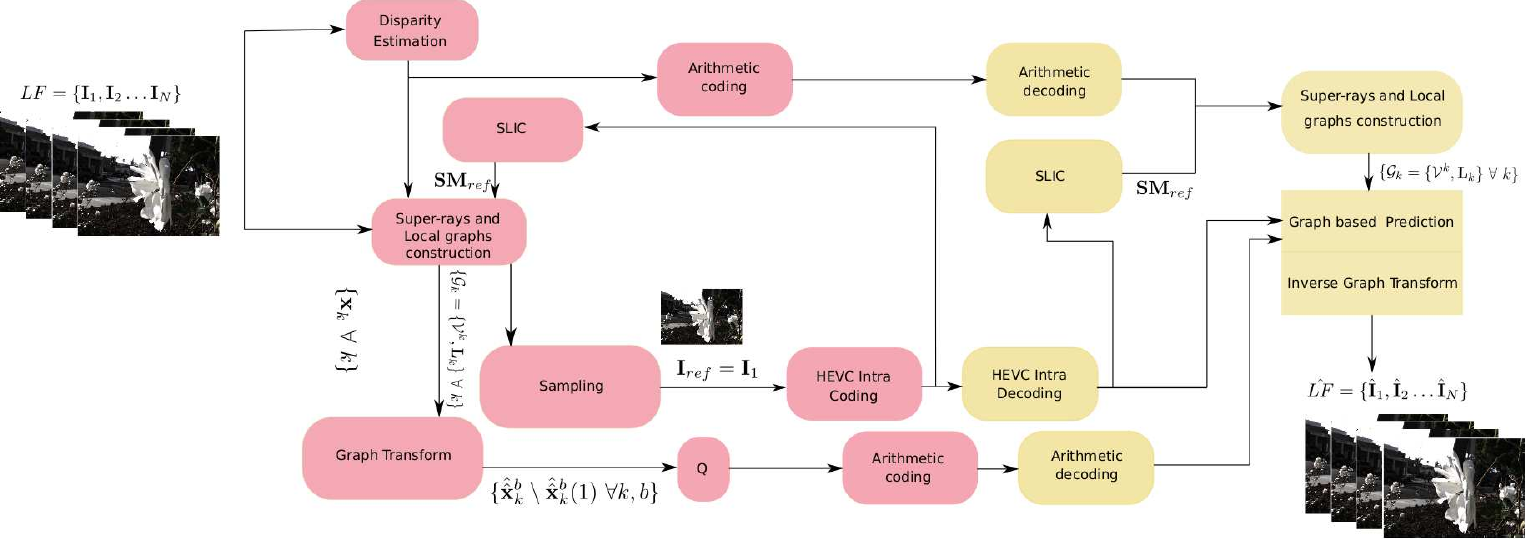}\\
\caption{Overview of proposed coding scheme with the separable graph transform.}
\label{fig:CodingSchemeSeparable}
\end{figure*}

The major difference between the coding scheme in the separable case (see Fig.\ref{fig:CodingSchemeSeparable}), compared with the non separable case, resides in the fact that the set of reference samples coded using HEVC-Intra is the top-left view. From this reference view, and the corresponding disparity map, that are transmitted, the decoder can compute the segmentation into super-pixels using the SLIC algorithm and then derive the super-rays used constructing the local graphs and compute the corresponding local graph transforms.

The spatio-angular high frequency graph transform coefficients are coded as in the previous scheme. Using the received reference view and the high frequency coefficients, the decoder 
can reconstruct all the views as explained in Section \ref{sec:GBSAP}. 

\section{Experimental analysis and results}
\begin{figure*}[!htp]
\centering
\includegraphics[width=0.23\linewidth]{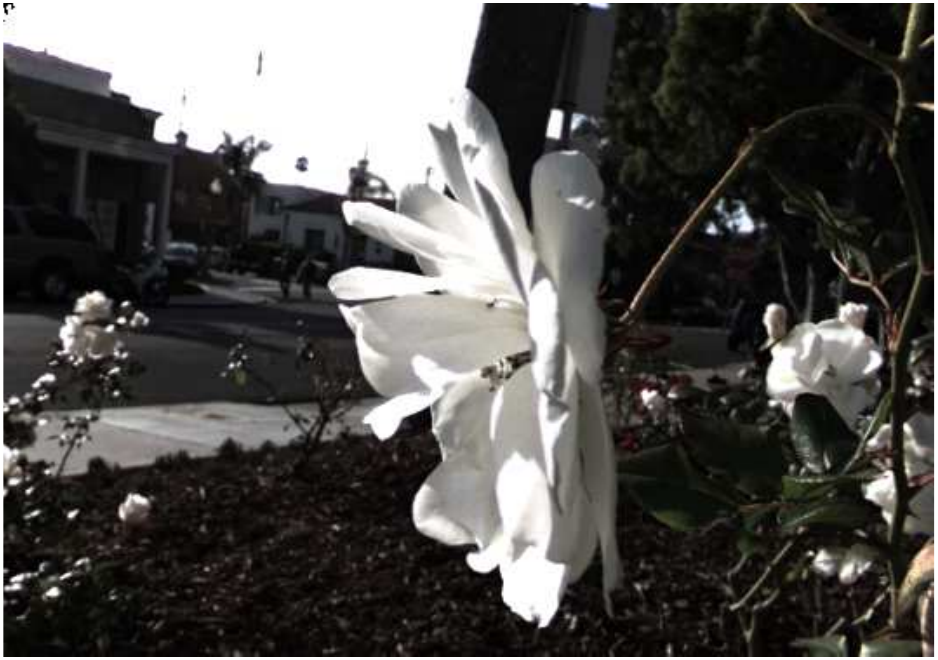}
\includegraphics[width=0.23\linewidth]{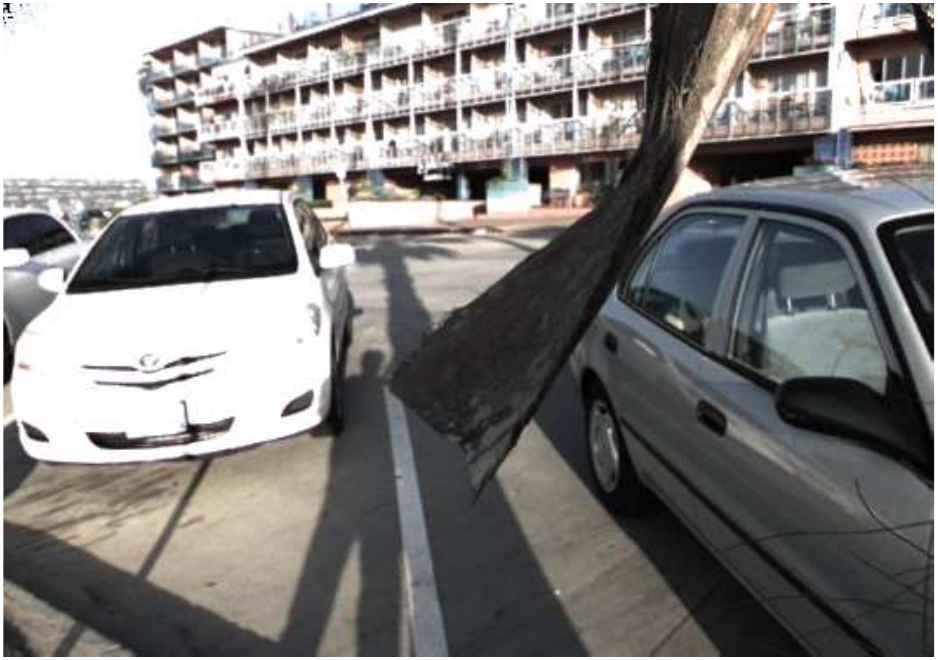}
\includegraphics[width=0.23\linewidth]{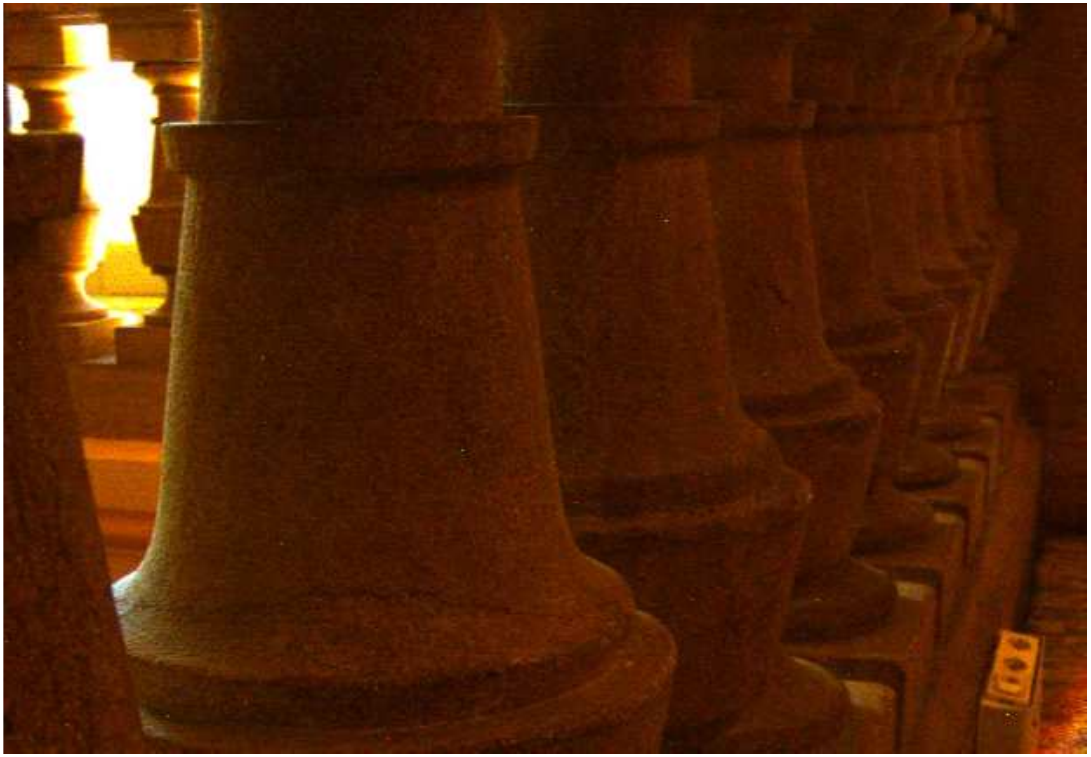}
\includegraphics[width=0.23\linewidth]{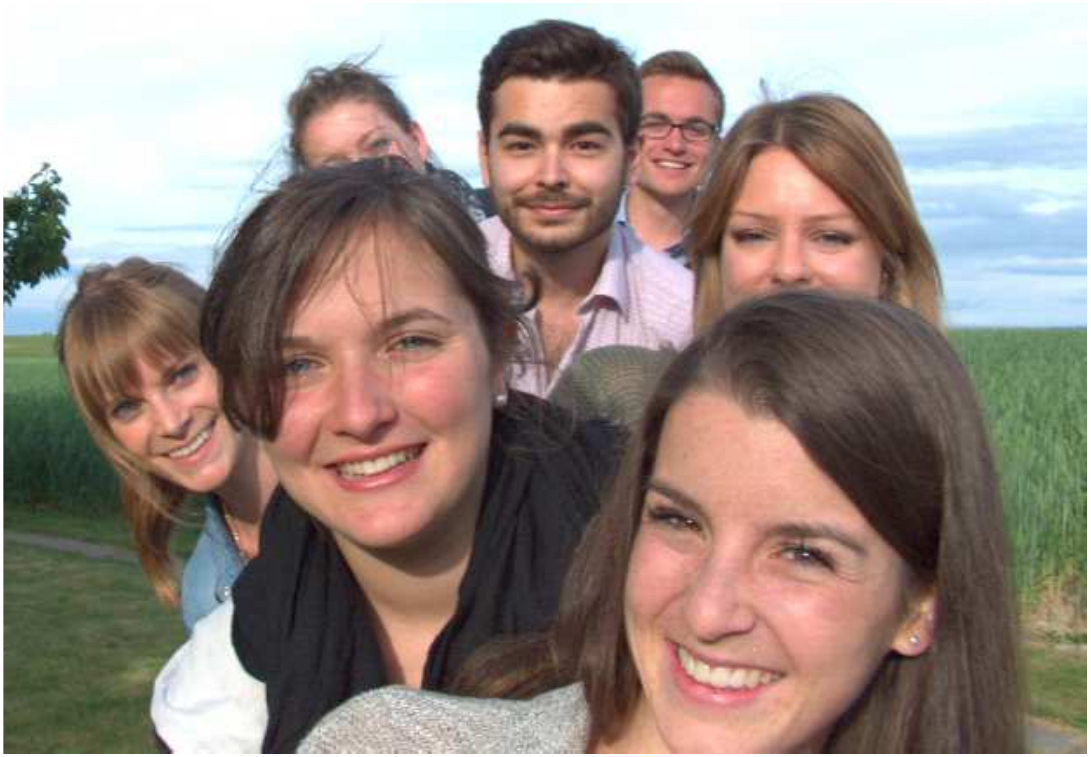}
\caption{Examples of light fields used in our experiments. Only the top-left view is shown for illustration purpose. From left to right: \textit{Flower2}, \textit{Cars}, \textit{StonePillarsInside} and \textit{Friends}.}
\label{fig:LF_test}
\end{figure*}

\begin{figure*}[!htp]
\centering
\includegraphics[width=0.23\linewidth]{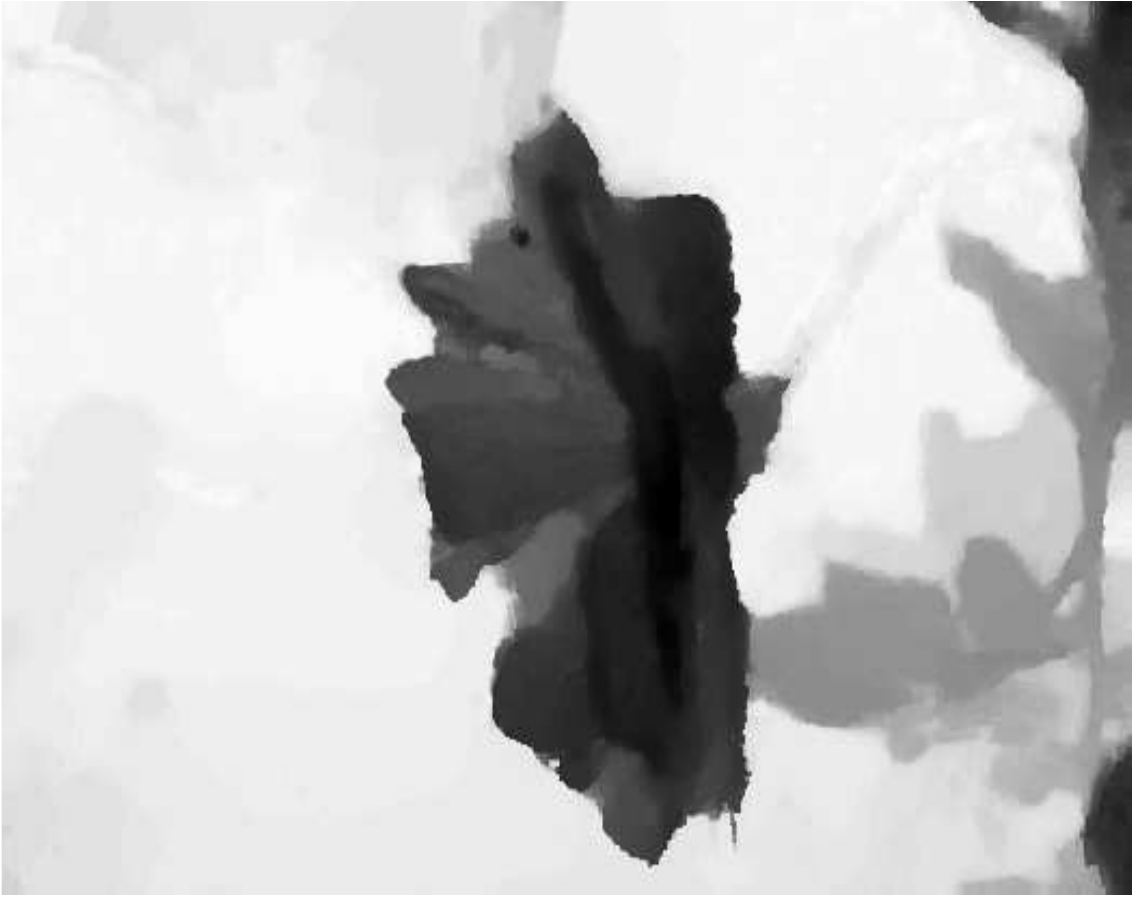} 
\includegraphics[width=0.23\linewidth]{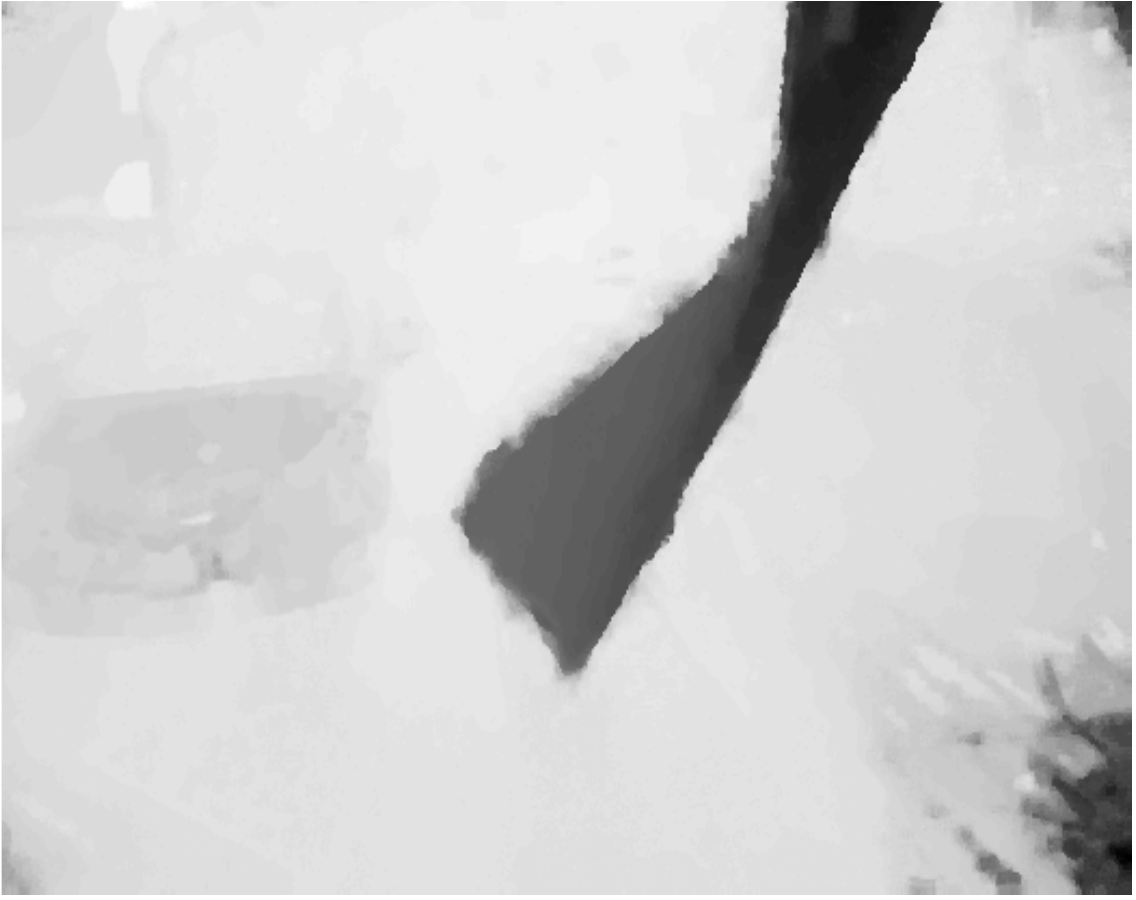}
\includegraphics[width=0.23\linewidth]{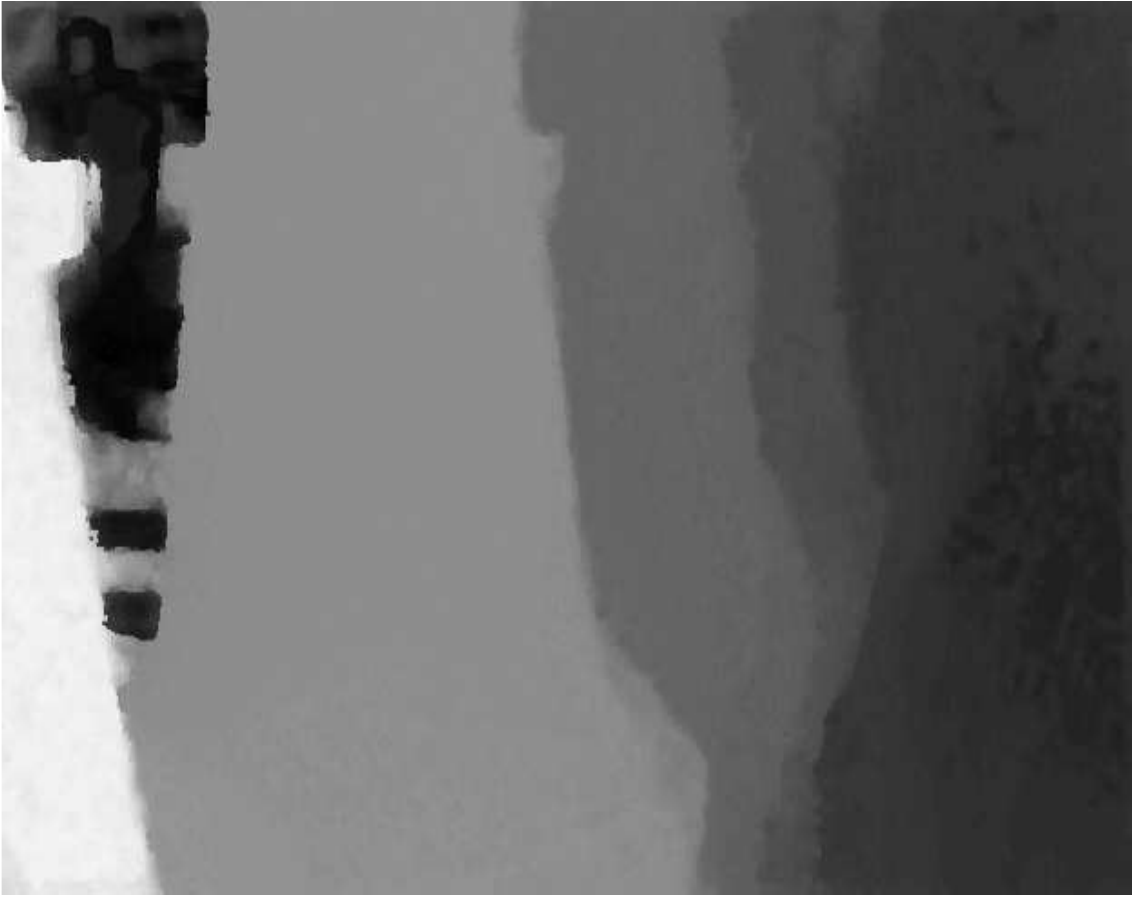}
\includegraphics[width=0.23\linewidth]{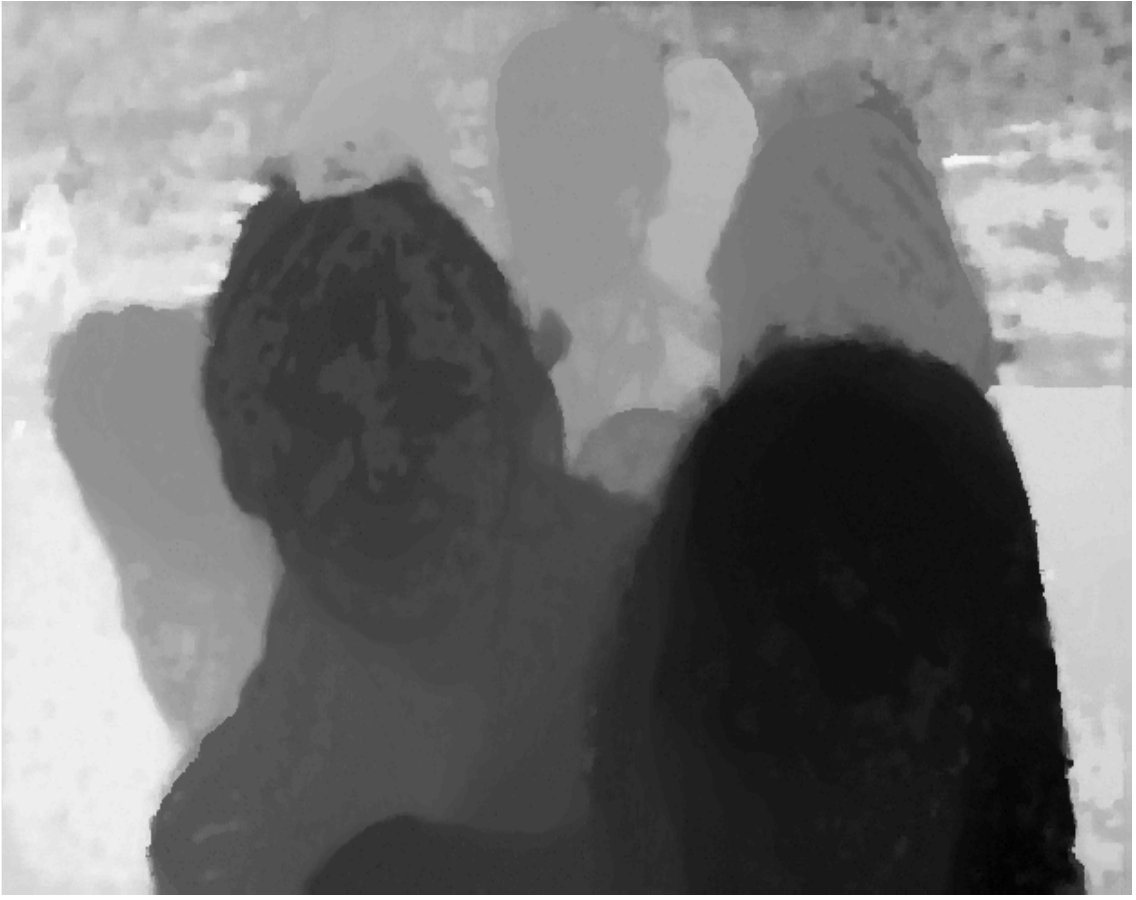}
\caption{Disparity maps for examples of light fields used in our experiments.}
\label{fig:LF_test_disp}
\end{figure*}
We applied both coding schemes on real light fields captured by plenoptic cameras from the datasets used in \cite{kalantari2016learning} and \cite{viola2018graph}. To avoid the strong vignetting and distortion problems on the views at the periphery of the light field, we only consider the $8 \times 8$ central sub-aperture images cropped to  $364 \times 524$ in \cite{kalantari2016learning}, and $9 \times 9$ cropped to $432 \times 624$ from \cite{viola2018graph}. 
Some of the light fields considered are shown in Figure \ref{fig:LF_test}. The full set of light fields considered for the test is: \textit{Flower2}, \textit{Cars}, \textit{Rock} and \textit{Seahorse} from the dataset in \cite{kalantari2016learning} and \textit{StonePillarInside} and \textit{Friends} from the dataset of ICIP challenge 2017 and used in \cite{viola2018graph}.
The method used to estimate the disparity of the top-left views is described in \cite{jiang2018depth}. Examples of disparity maps provided are shown in Fig. \ref{fig:LF_test_disp}. A sparse set of disparity values and the segmentation map of the reference view are computed with SLIC \cite{achanta2012slic}, and used to construct super-rays, i.e., the local graph supports as described in Section \ref{sec:GraphConstruction}. We set the number of super-rays to $4000$ to cope with the complexity issue of the non separable graph transform and show the effect of our graph prediction (the importance of exploring the long term dependencies between super-rays). 

\subsection{Non Separable vs Separable Graph Prediction}

\subsubsection{Energy compaction}
As explained before, we aim at compacting most of the light field energy in few coefficients, and at then predicting these coefficients (i.e. they are not transmitted) from a coded reference image and from the high frequency graph transform coefficients that need to be transmitted but at small cost given that they contain little information. Table \ref{tab:percentage_energy_predicted} gives the percentage of total energy that resides in the predicted DC spatio-angular bands for both non separable ($\hat{\mathbf{x}}_k(\mathcal{T}) \ \forall \ k$) and separable ($\doublehat{\mathbf{x}}^b_k(1) \ \forall k,b$) cases. 
We can observe that most of the energy is compacted in the DC spatio-angular bands, which shows the efficiency in terms of spatio-angular de-correlation of the graph transforms.

The non separable prediction has the benefit of the low energy of the high frequency coefficients of the graph transform that also need to be coded. The separable graph transform, in some cases, looses this benefit as we predict the DC angular(i.e. after the transform across the views) coefficients of all spatial bands. Those low angular frequency coefficients may not contain all the energy otherwise captured by the lower spatio-angular frequency coefficients of the non separable case, although it remains quite efficient in terms of energy compaction as we can see in Table \ref{tab:percentage_energy_predicted}. 

\begin{figure*}[htp!]
	\centering
	\includegraphics[width=0.6\linewidth]{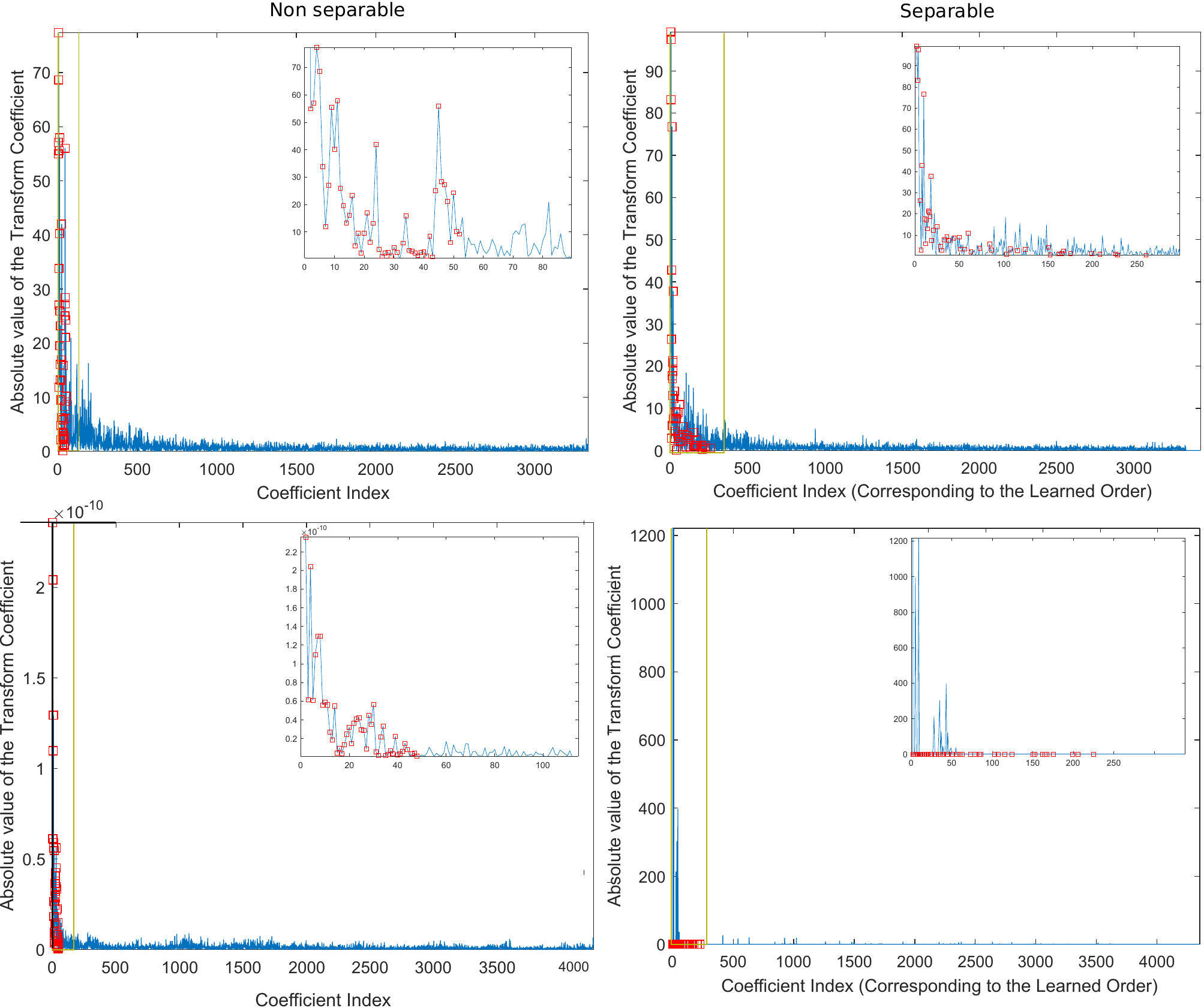}\\
	\caption{Energy compaction for two super-rays of  \textit{Flower2}. The transform coefficients are ordered with the learned frequency order. The red squares are the predicted DC values on the decoder side. The two rows correspond to two different super-rays and the two columns are for both cases: Non Separable and Separable respectively.}
	\label{fig:illustration_COMPACT}
\end{figure*}

To further illustrate the energy compaction of the transforms, we plot in Fig. \ref{fig:illustration_COMPACT}, for two different super-rays, the transform coefficients (except the first one which holds most of the energy) following the coding order (learned order of frequencies) for both cases: separable and non separable graph transforms. As we can see, in the non separable case, the low frequencies that are predicted on the decoder side (the red dots) correspond to the first frequencies and thus to those who hold most of the energy. However, in the separable case, the coefficients predicted do not necessarily exhibit the highest energy. This is quite clear in the second example, where the red dots in the separable case are assigned to very low values, and some high values still need to be sent to the decoder side. For the non separable case, the decorrelation has been perfect, and there is a very low energy in the coefficients shown in the plot.

\subsubsection{Compressibility of the reference view }
Thanks to the prediction equations introduced in Section \ref{sec:GBSAP}, an efficient encoding of the top-left view in the separable case or the reference view in the non separable case (using any classical encoder with efficient spatial predictors) can be seen as a way to encode those DC spatio-angular frequency coefficients which contain most of the light field energy.

The separable graph transform based prediction takes advantage of the natural structure of the reference view as we can see in Fig. \ref{fig:samples_example}. It is thus efficiently coded using intra-prediction tools. For the non-separable graph prediction, however, this is not the case since the optimal sampling does not totally guarantee that the samples are well structured in each super-ray of the 2D reference view. Yet, the super-ray segmentation preserves in a certain way the natural structure of the reference view. 

In our experiments, we choose HEVC intra to encode this information (i.e. top left view or reference view). Tables \ref{tab:ratecompare_HEVCentropy} and \ref{tab:ratecompare_HEVCentropy_} give the bit rate obtained when encoding the reference view (from which are derived the DC spatio-angular frequency coefficients) with HEVC-Intra (with QP set to $0$). The bit rates are compared with those  obtained when using a simple arithmetic coder for directly encoding the spatio-angular DC coefficients. In order to apply the arithmetic coder,
for each frequency band $b$, we first group all the coefficients $\doublehat{\mathbf{x}}^b_k(1) \ \forall k$ of the super-rays in which this band exists, and we code them with an arithmetic coder independently of the other bands.
The table shows the rate gain obtained by encoding the set of reference samples with HEVC intra, thanks to the possibility to capture dependencies between super-rays.

\begin{table*}[ht]
	\centering
	\begin{tabular}{c||c|c}
		Light Fields & Energy Percentage in $\hat{x}_k(\mathcal{T}) \ \forall \ k$ & Energy Percentage in $\doublehat{\mathbf{x}}^{b}_{k}(1) \ \forall \ k,b$ \\
		\hline
		Flower 2 & 99.15 \% & 99.02 \%  \\
		Cars & 99.27 \% &  99.34 \% \\
		Rock & 98.63 \% & 98.45 \% \\
		Seahorse & 99.17 \% & 98.73 \% \\  
		Stone Pillars Inside & 98.90 \% & 98.26 \%  \\
		Friends & 99.76 \% & 99.80 \% 
	\end{tabular}
		\caption{Percentage of energy residing in the DC spatio-angular bands $\hat{x}_k(\mathcal{T}) \ \forall \ k$ in the non separable case (left column), and in the bands $\doublehat{\mathbf{x}}^{b}_{k}(1) \ \forall \ k,b$ in the separable case (right column).}
		\label{tab:percentage_energy_predicted}
\end{table*}


\begin{table*}[!h]
\caption{Bit rate obtained, in the case of the non separable graph transform, when using HEVC intra-coding (left column) of the reference set of samples, and entropy (arithmetic) coding of all the DC spatio-angular bands $\hat{\mathbf{x}}_{k}(\mathcal{T}) \ \forall \ k$ (right column).}
\label{tab:ratecompare_HEVCentropy}
\centering
\begin{tabular}{c||c|c}
	Light Fields & HEVC-Intra coding of set of reference samples & Entropy coding of $\hat{\mathbf{x}}_{k}(\mathcal{T}) \forall k$ \\
	\hline
	Flower 2 & 1.23 Mbits & 1.59 Mbits \\
	Cars & 1.45 Mbits & 1.64 Mbits \\
	Rock & 1.31 Mbits & 1.49 Mbits\\
	Seahorse & 0.74 Mbits & 1.38 Mbits \\  
	Stone Pillars Inside & 1.92 Mbits & 1.84 Mbits  \\
	Friends & 1.86 Mbits & 2.07 Mbits
\end{tabular}
\end{table*}

\begin{table*}[h]
	\caption{Bit rate obtained, in the case of the separable graph transform, when using HEVC intra to code the first view (left column), and when using entropy (arithmetic) coding of all the DC spatio-angular bands $\doublehat{\mathbf{x}}^{b}_{k}(1) \ \forall \ k,b$ (right column).}
	\label{tab:ratecompare_HEVCentropy_}
	\centering
	\begin{tabular}{c||c|c}
		Light Fields & HEVC-Intra of reference view & Entropy coding of $\doublehat{\mathbf{x}}^{b}_{k}(1) \ \forall \ k,b$ \\
		\hline
		Flower 2 & 1.0 Mbits & 1.48 Mbits \\
		Cars & 1.06 Mbits & 1.54 Mbits \\
		Rock & 1.02 Mbits & 1.38 Mbits\\
		Seahorse & 0.72 Mbits & 1.22 Mbits \\  
		Stone Pillars Inside & 1.81 Mbits & 1.66 Mbits  \\
		Friends & 1.62 Mbits & 2.04 Mbits
\end{tabular}
\end{table*}

\subsubsection{Robustness of the Prediction}
\begin{figure}[ht!]
	\centering
	\includegraphics[width=0.8\textwidth]{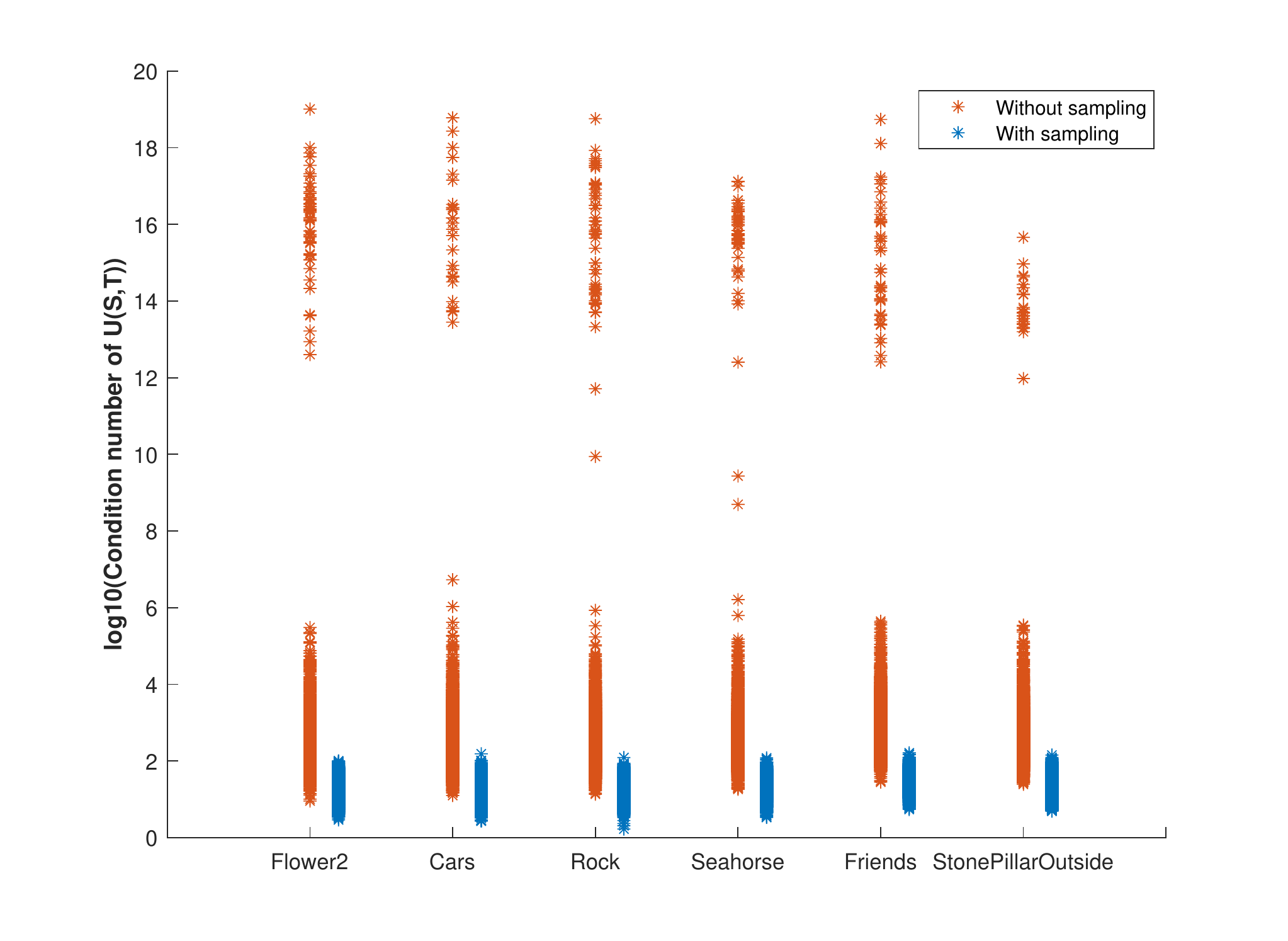}\\
	\caption{Efficiency of the sampling and effect on the condition number of the matrix $\mathbf{U}_k(\mathcal{S},\mathcal{T})$. We show for each dataset, for all super-rays the log base 10 of the condition number without (red) and with (blue) sampling.}
	\label{fig:conditioning}
\end{figure}

In order to assess the efficiency of our prediction and the light field sampling algorithm for the non separable case, we plot in Fig. \ref{fig:conditioning} the condition number in log base $10$ of the matrix $\mathbf{U}_k(\mathcal{S},\mathcal{T})$ for all super-rays $k$ in all the datasets. The condition number is measured to show how much sensible is our prediction in equation \ref{eq:predictionNS} $\hat{\mathbf{x}}_k(\mathcal{T})$ to a small change in $\Big(\mathbf{x}_k(\mathcal{S}) -  \mathbf{U}_k(\mathcal{S},\mathcal{T}_c) \hat{\mathbf{x}}_k(\mathcal{T}_c) \Big)$. On one hand, the condition numbers are computed without sampling i.e. assuming that the reference samples are those in the top-left view. These are shown in red, while the results in blue correspond to the condition numbers after taking the actual samples found with algorithm \ref{alg:SRsampling}. A major difference is shown in log scale, where the sampling has reduced the condition number from $10^{15}$ to a maximum of around $10^2$. Without sampling, the prediction fails since a tiny change in the high frequency coefficients (even a small rounding procedure) can result in a huge loss in the reconstruction quality. 

For the prediction based on the separable graph transform, we do not need a matrix inversion. We only need to invert a number $\mathbf{V}_k^b(1,1)$ whose minimum corresponds to $1/\sqrt{M \times N}$. This inversion does have a smaller impact than the one in the non separable case. This is a major explanation of the PSNR difference between our schemes for a fixed quantization step size $Q=1$ in Table \ref{tab:ratecompare}.

\subsection{RD Comparison against state of the art coders} 
We assess the proposed graph-based spatio-angular prediction methods in the context of quasi-lossless light field coding in comparison with a complete HEVC-based scheme with a QP set to $0$ and a GOP size of $4$. The HEVC version used in the tests is the HM-16.10. The light fields are coded following a raster scan order starting with the top-left view as a reference Intra-coded frame. Results are reported in Table \ref{tab:ratecompare} where we compare mainly the bit rate needed to code a light field in a quasi-lossless setting (we consider a PSNR higher that $50$ dB as a quasi-lossless compression). A substantial gain in bit rate is observed while preserving a high quality of the reconstructed light fields. This can be justified by the efficiency of the proposed spatio-angular graph transforms in terms of energy compaction along with the ability of HEVC-intra to effectively exploit spatial correlation in the reference view.

\begin{table*}[!h]
\caption{Rate comparison between our proposed schemes (with both non separable and separable graph transforms) and a scheme using HEVC-inter to code the views in a raster scan order, at high quality (PSNR $>50$ dB)}
\label{tab:ratecompare}
\centering
\resizebox{\linewidth}{!}{%
\begin{tabular}{c||c|c|c|c}
	Light Fields & HEVC-Inter (QP=0) Raster Scan & Non Separable Scheme (Q = 0.5) & Non Separable Scheme (Q = 1) & Separable Scheme (Q = 1) \\
	\hline
	Flower 2 & 3.3129 bpp (54.2033 dB) & 2.4470 bpp (60.4656 dB) & 2.4457 bpp (52.9393 dB) & 2.4799 bpp (55.1969 dB)\\
	Cars & 3.6688 bpp (54.0812 dB) & 2.7759 bpp (60.5035 dB)  & 2.7801 bpp (53.0268 dB) & 2.6258 bpp (55.2009 dB) \\
	Rock & 3.2700 bpp (53.7601 dB) &  2.0423 bpp (60.2994 dB) & 2.0545 bpp (52.6230 dB) & 2.0162 bpp (54.7765 dB)\\
	Seahorse & 2.4751 bpp (54.3804 db) & 1.8224 bpp (60.4474 dB) & 1.7849 bpp (53.0111 dB) & 1.9762 bpp (55.2844 dB) \\  
	Stone Pillars Inside & 4.9017 bpp (52.1036 dB) & 2.5559 bpp (59.7134 dB) & 1.5269 bpp (52.3953 dB) & 3.3094 bpp (55.0022 dB)\\
	Friends & 3.5400 bpp (52.7986 dB)  & 1.9327 bpp (59.7657 dB) & 1.9311 bpp (52.4402 dB) & 2.4436 bpp (54.8196 dB)
\end{tabular}}
\end{table*}


\section{Conclusion}
In this paper, we have proposed sampling and prediction methods with local graph transforms for light field energy compaction and compression.
Based on the graph sampling theory, the proposed methods allow taking advantage of the good energy compaction property of the graph transform on local supports with a limited complexity, while benefiting from well established but powerful prediction mechanisms in the pixel domain. We considered both a super-ray based non separable graph transform and a spatio-angular separable simplified version. Two coding schemes have been described based on the non separable and separable graph transforms. The schemes have been 
assessed for high quality (quasi-lossless) coding. Both proposed approaches are very efficient when the quantization noise on the reference set of samples is low, hence for quasi-lossless compression. If the reference set of samples is too coarsely quantized, drift and noise amplification may appear during the prediction step. This is due to the fact that, in Equation (\ref{eq:prediction}), the prediction uses the spatial transform coefficients estimated on the reference set of samples available at the decoder side. Further study will be dedicated to addressing this problem in the case of lossy compression.

\bibliographystyle{IEEEtran}
\bibliography{biblio}
%








\end{document}